\documentclass{llncs} 
\usepackage{hyperref}
\usepackage{url}
\usepackage{amsmath}
\usepackage{amssymb}
\usepackage[ruled]{algorithm}
\usepackage{algpseudocode}
\usepackage{xspace}
\usepackage{caption}
\usepackage{subcaption}
\usepackage{multirow}
\usepackage{graphicx}

\newcommand{\randn}{\operatorname{randn}}
\newcommand{\orth}{\operatorname{orthogonalize}}
\newcommand{\eig}{\operatorname{eig}}

\newcommand{\R}{\mathbb{R}}
\newcommand{\rank}{\operatorname{rank}}
\DeclareMathOperator*{\argmin}{arg\,min}
\newcommand\defeq{\mathrel{\overset{\makebox[0pt]{\mbox{\normalfont\tiny\sffamily def}}}{=}}}
\newcommand{\RandEmbed}{Rembrandt\xspace}
\newcommand{\darkknowledge}{\mathbb{E}[\mathbb{E}[y | x]^\top \mathbb{E}[y | x]]}

\newcommand{\citep}[1]{\cite{#1}}

\begin{document}

\author{
Paul Mineiro \and Nikos Karampatziakis
}
\institute{Microsoft Cloud Information Services Lab \\
\email{\{pmineiro,nikosk\}@microsoft.com} 
}

\title{Fast Label Embeddings via Randomized Linear Algebra}

\maketitle

\begin{abstract}
Many modern multiclass and multilabel problems are characterized
by increasingly large output spaces.   For these problems, label
embeddings have been shown to be a useful primitive that can improve
computational and statistical efficiency.  In this work we utilize a
correspondence between rank constrained estimation and low dimensional
label embeddings that uncovers a fast label embedding algorithm which
works in both the multiclass and multilabel settings.  The result is a
randomized algorithm whose running time is exponentially faster than
naive algorithms.  We demonstrate our techniques on two large-scale
public datasets, from the Large Scale Hierarchical Text Challenge and
the Open Directory Project, where we obtain state of the art results.
\end{abstract}

\section{Introduction}

Recent years have witnessed the emergence of many multiclass and
multilabel datasets with increasing number of possible labels, such as
ImageNet~\citep{deng2009imagenet} and the Large Scale Hierarchical Text
Classification (LSHTC) datasets~\citep{kosmopoulos2010ecir}.  One could
argue that all problems of vision and language in the wild have extremely
large output spaces.

When the number of possible outputs is modest, multiclass and multilabel
problems can be dealt with directly (via a max or softmax layer) or with
a reduction to binary classification.  However, when the output space is
large, these strategies are too generic and do not fully exploit some
of the common properties that these problems exhibit. For example,
often the alternatives in the output space have varying degrees of
similarity between them so that typical examples from similar classes
tend to be closer\footnote{or more confusable, by machines and humans
alike.} to each other than from dissimilar classes. More concretely,
classifying an image of a Labrador retriever as a golden retriever is
a more benign mistake than classifying it as a rowboat.

Shouldn't these problems then be studied as structured prediction
problems, where an algorithm can take advantage of the structure?  That
would be the case if for every problem there was an unequivocal structure
(e.g. a hierarchy) that everyone agreed on and that structure
was designed with the goal of being beneficial to a classifier. When
this is not the case, we can instead let the algorithm uncover
a structure that matches its own capabilities.

In this paper we use label embeddings as the underlying structure
that can help us tackle problems with large output spaces, also known
as extreme classification problems. Label embeddings can offer improved
computational efficiency because the embedding dimension is much smaller
than the dimension of the output space. If designed carefully and applied
judiciously, embeddings can also offer statistical efficiency because
the number of parameters can be greatly reduced without increasing,
or even reducing, generalization error.

\subsection{Contributions}

We motivate a particular label embedding defined by the low-rank
approximation of a particular matrix, based upon a correspondence
between label embedding and the optimal rank-constrained least squares
estimator.  Assuming realizability and infinite data, the matrix being
decomposed is the expected outer product of the conditional label
probabilities.  In particular, this indicates two labels are similar
when their conditional probabilities are linearly dependent across the
dataset. This unifies prior work utilizing the confusion matrix for
multiclass~\citep{bengio2010label} and the empirical label covariance
for multilabel~\citep{tai2012multilabel}.

We apply techniques from randomized linear
algebra~\citep{halko2011finding} to develop an efficient and scalable
algorithm for constructing the embeddings, essentially via a novel
randomized algorithm for rank-constrained squared loss regression.
Intuitively, this technique implicitly decomposes the prediction matrix
of a model which would be prohibitively expensive to form explicitly. The
first step of our algorithm resembles compressed sensing approaches to
extreme classification that use random matrices~\citep{hsu2009multi}.
However our subsequent steps tune the embeddings to the data at hand,
providing the opportunity for empirical superiority.

\section{Algorithm Derivation} \label{sec:algorithm}

\subsection{Notation}

We denote vectors by lowercase letters $x$, $y$ etc.\ and matrices by
uppercase letters $W$, $Z$ etc.  The input dimension is denoted by $d$,
the output dimension by $c$ and the embedding dimension by $k$. For
multiclass problems $y$ is a one hot (row) vector (i.e.~a vertex of the
$c-1$ unit simplex) while for multilabel problems $y$ is a binary vector
(i.e.~a vertex of the unit $c$-cube).  For an $m\times n$ matrix $X \in
\R^{m \times n}$ we use $||X||_F$ for its Frobenius norm, $X^\dagger$
for the pseudoinverse, $\Pi_{X,L}$ for the projection onto the left
singular subspace of $X$, and $X_{1:k}$ for the matrix resulting by
taking the first $k$ columns of $X$.  We use $X^*$ to denote a matrix
obtained by solving an optimization problem over matrix parameter $X$.
The expectation of a random variable $v$ is denoted by $\mathbb{E}[v]$.

\subsection{Background}

\begin{algorithm}[t]
  \caption{Randomized PCA}
  \begin{algorithmic}[1]
    \Statex
    \Function{RPCA}{$k, X \in \R^{n \times d}$}
      \State $(p, q) \leftarrow (20, 1)$ \Comment{These hyperparameters rarely need adjustment.}
      \State $Q \leftarrow \randn(d,k+p)$
      \For{$i \in \{ 1, \ldots, q \}$} \Comment{Randomized range finder for $X^\top X$}
            \State $\Psi \leftarrow X^\top X Q$ \Comment{$\Psi$ can be computed in one pass over the data}
	        \State $Q \leftarrow \orth(\Psi)$ \Comment{orth complexity $O(dk^2)$ is independent of $n$}
      \EndFor \label{lin:pcarangefindend} \Comment{NB: total of $(q+1)$ data passes, including next line}
      \State $F \leftarrow (X^\top X Q)^\top (X^\top X Q)$ \label{lin:pcafinalpass} \Comment{$F \in \R^{(k+p)\times(k+p)}$ is ``small'}
      \State $(V, \Sigma^2) \leftarrow \eig(F, k)$ \Comment{Exact optimization on small matrix}
      \State $V \leftarrow (X^\top X Q) V \Sigma^\dagger$  \Comment{Back out the solution}
      \State \Return $(V, \Sigma)$
    \EndFunction
  \end{algorithmic}
  \label{alg:rpca}
\end{algorithm}

In this section we offer an informal discussion of randomized algorithms
for approximating the principal components analysis of a data matrix
$X \in \R^{n\times d}$ with $n$ examples and $d$ features. For a very
thorough and more formal discussion see \cite{halko2011finding}.

Algorithm~\ref{alg:rpca} shows a recipe for performing randomized PCA.  In
both theory and practice, the algorithm is insensitive to the parameters
$p$ and $q$ as long as they are large enough (in our experiments we use
$p=20$ and $q=1$). We start with a set of $k+p$ random vectors and use
them to probe the range of $X^\top X$. Since principal eigenvectors can
be thought as ``frequent directions'' \citep{liberty2013simple}, the
range of $\Psi$ will tend to be more aligned with the space spanned by
the top eigenvectors of $X^\top X$. We compute an orthogonal basis for
the range of $\Psi$ and repeat the process $q$ times. This can also be
thought as orthogonal (aka subspace) iteration for finding eigenvectors
with the caveat that we early stop (i.e., $q$ is small). Once we are done
and we have a good approximation for the principal subspace of $X^\top X$,
we optimize fully over that subspace and back out the solution. The last
few steps are cheap because we are only working with a $(k+p)\times(k+p)$
matrix and the largest bottleneck is either the computation of $\Psi$ in
a single machine setting or the orthogonalization step if parallelization
is employed. An important observation we use below is that $X$ or $X^\top
X$ need not be available explicitly; to run the algorithm we only need
to be able to compute the result of multiplying with $X^\top X$.

\subsection{Rank-Constrained Estimation and Embedding}

\begin{algorithm}[t]
  \caption{Rembrandt: Response EMBedding via RANDomized Techniques.  A reference implementation is publicly available.\cite{mineirogit2015}}
  \begin{algorithmic}[1]
    \Statex
    \Function{Rembrandt}{$k, X \in \R^{n \times d}, Y \in \R^{n \times c}$}
      \State $(p, q) \leftarrow (20, 1)$ \Comment{These hyperparameters rarely need adjustment.}
      \State $Q \leftarrow \randn(c, k+p)$  \label{lin:rangefindstart}
      \For{$i \in \{ 1, \ldots, q \}$} \Comment{Randomized range finder for $Y^\top \Pi_{X,L} Y$}
        \State $Z \leftarrow \argmin \| YQ - XZ \|^2_F$ \label{lin:learningstep}
        \State $Q \leftarrow \orth(Y^\top X Z)$
      \EndFor \label{lin:rangefindend} \Comment{NB: total of $(q+1)$ data passes, including next line}
      \State $F \leftarrow (Y^\top X Z)^\top (Y^\top X Z)$ \Comment{$F \in \R^{(k+p)\times(k+p)}$ is ``small''}
      \State $(V, \Sigma^2) \leftarrow \eig(F, k)$
      \State $V \leftarrow (Y^\top X Z) V \Sigma^\dagger$ \Comment{$V \in \R^{c \times k}$ is the embedding}
      \State \Return $(V, \Sigma)$
    \EndFunction
  \end{algorithmic}
  \label{alg:rembed}
\end{algorithm}

We begin with a setting superficially unrelated to label embedding.
Suppose we seek an optimal squared loss predictor of a high-cardinality
target vector $y \in \R^c$ which is linear in a high dimensional feature
vector $x \in \R^d$.  Due to sample complexity concerns, we impose a
low-rank constraint on the weight matrix.  In matrix form,
\begin{align}
W^* &= \argmin_{W \in \R^{d \times c} | \rank(W) \leq k} \| Y - X W \|^2_F, \label{eqn:rankconstrained}
\end{align}
where $Y \in \R^{n \times c}$ and $X \in \R^{n \times d}$ are the
target and design matrices respectively.  This is a special case of
a more general problem studied by \cite{friedland2007generalized};
specializing their result yields the solution $W^* = X^\dagger (\Pi_{X,L}
Y)_k$, where $\Pi_{X,L}$ projects onto the left singular subspace of $X$,
and $(\cdot)_k$ denotes optimal Frobenius norm rank-$k$ approximation,
which can be computed\footnote{if~$X=U_X\Sigma_X V_X^\top$ is the SVD of
$X$, then $\Pi_{X,L}=U_XU_X^\top$.} via SVD.  The expression for $W^*$
can be written in terms of the SVD $\Pi_{X,L} Y = U \Sigma V^\top$,
which, after simple algebra, yields $W^* = \left( X^\dagger (Y V_{1:k})
\right) V_{1:k}^\top$.  This is equivalent to the following procedure:
\begin{enumerate}
\item $YV_{1:k}$: Project $Y$ down to $k$ dimensions using the top right
singular vectors of $\Pi_{X,L} Y$.
\item $X^\dagger (Y V_{1:k})$ Least squares fit the projected labels
using $X$ and predict them.
\item $ \left( X^\dagger (Y V_{1:k}) \right) V_{1:k}^\top$: Map
predictions to the original output space, using the transpose of the
top right singular vectors of $\Pi_{X,L} Y$.
\end{enumerate}
This motivates the use of the right singular vectors of $\Pi_{X,L} Y$ as a
label embedding.  The $\Pi_{X,L} Y$ term can be demystified: 
it corresponds to the predictions of the optimal unconstrained model, \[
\begin{aligned}
Z^* &= \argmin_{Z \in \R^{d \times c}} \| Y - X Z \|^2_F, \\
\Pi_{X,L} Y &= X Z^* \defeq \hat Y.
\end{aligned}
\]  The right singular vectors $V$ of $\Pi_{X,L} Y$ are therefore the
eigenvectors of $\hat Y^\top \hat Y $, i.e., the matrix formed by the
sum of outer products of the optimal unconstrained model's predictions
on each example.  Note that actually computing and materializing 
$Z^* \in \R^{d \times c}$ would be expensive; a key aspect of 
the randomized algorithm is that we get the same result while
avoiding this intermediate.  In particular we can find the product of
$\Pi_{X,L} Y$ with another matrix $Q \in \R^{c \times k}$ via 
\begin{align}
Z^* Q &= \argmin_{Z \in \R^{d \times k}} \| Y Q - X Z \|^2_F, \nonumber \\
\Pi_{X,L} Y Q &= X Z^* Q. \label{eqn:multop}
\end{align}
Because squared loss is a proper scoring rule it is minimized at the
conditional mean.  In the limit of infinite training data ($n\to \infty$)
and sufficient model flexibility (so that $\hat{y} = \mathbb{E}[y|x]$)
we have that
\begin{equation}
\frac{1}{n} \hat Y^\top \hat Y  \stackrel{\textrm{a.s.}}{\longrightarrow} 
\darkknowledge \label{eqn:darkknowledge}
\end{equation}
by the strong law of large numbers. An embedding based upon the
eigendecomposition of $\darkknowledge$ is not practically actionable, but
does provide valuable insights.  For example, the principal label space
transformation of \cite{tai2012multilabel} is an eigendecomposition of
the empirical label covariance $Y^\top Y$.  This is a plausible
approximation to $\darkknowledge$ in the multilabel case.  However,
for multiclass (or multilabel where most examples have at most one
nonzero component), the low-rank constraint alone cannot produce good
generalization if the input representation is sufficiently flexible;
the eigendecomposition of the prediction covariance will merely select
a basis for the $k$ most frequent labels due to the absence of empirical
cooccurence statistics.  Under these conditions we must further regularize
(i.e., tradeoff variance for bias) beyond the low-rank constraint,
so that $\hat Y$ better approximates $\mathbb{E}[Y|X]$ rather than the
observed $Y$.  Our procedure admits tuning the bias-variance tradeoff
via choice of model (features) used in line \ref{lin:learningstep}
of Algorithm~\ref{alg:rembed}.

\subsection{\RandEmbed}

Our proposal is \RandEmbed, described in Algorithm \ref{alg:rembed}.
In the previous section, we motivated the use of the top right singular
space of $\Pi_{X,L} Y$ as a label embedding, or equivalently, the top
principal components of $Y^\top \Pi_{X,L} Y$ (leveraging the fact
that the projection is idempotent).  Using randomized techniques,
we can decompose this matrix without explicitly forming it, because
we can compute the product of $\Pi_{X,L} Y$ with another matrix $Q$ via
equation \ref{eqn:multop}.  Algorithm \ref{alg:rembed} is a specialization
of randomized PCA to this particular form of the matrix multiplication
operator.  Starting from a random label embedding which satisfies the
conditions for randomized PCA (e.g., a Gaussian random matrix), the
algorithm first fits the embedding, outer products the embedding with
the labels, orthogonalizes and repeats for some number of iterations.
Then a final exact eigendecomposition is used to remove the additional
dimensions of the embedding that were added to improve convergence.
Note that the optimization of \ref{eqn:multop} is over $\R^{d \times
(k+p)}$, not $\R^{d \times c}$, although the result is equivalent;
this is the main computational advantage of our technique.

The connection to compressed sensing approaches to extreme classification
is now clear, as the random sensing matrix corresponds to the starting
point of the iterations in Algorithm \ref{alg:rembed}.  In other words,
compressed sensing corresponds to Algorithm \ref{alg:rembed} with $q=0$
and $p = 0$, which results in a whitened random projection of the labels
as the embedding.  Additional iterations ($q > 0$) and oversampling
($p > 0$) improve the approximation of the top eigenspace, hence the
potential for improved performance.  However when the model is
sufficiently flexible, an embedding matrix which ignores the
training data might be superior to one which overfits the training data.

Equation \eqref{eqn:multop} is inexpensive to compute.  The matrix vector
product $Y Q$ is a sparse matrix-vector product so complexity $O(n
s k)$ depends only on the average (label) sparsity per example $s$
and the embedding dimension $k$, and is independent of the number of
classes $c$.  The fit is done in the embedding space and therefore is
independent of the number of classes $c$, and the outer product with the
predicted embedding is again a sparse product with complexity $O(n s k)$.
The orthogonalization step is $O(ck^2)$, but this is amortized over the
data set and essentially irrelevant as long as $n > c$.  While random
projection theory suggests $k$ should grow logarithmically with $c$,
this is only a mild dependence on the number of classes.

\section{Related Work} \label{sec:relatedwork}

Low-dimensional dense embeddings of sparse high-cardinality output
spaces have been leveraged extensively in the literature, due to their
beneficial impact on multiple algorithmic desiderata.  As this work
emphasizes, there are potential statistical (i.e., regularization)
benefits to label embeddings, corresponding to the rich literature
of low-rank regression regularization~\citep{izenman1975reduced}.
Another common motivation is to mitigate space or time complexity
at training or inference time.  Finally, embeddings can be part of
a strategy for zero-shot learning~\citep{palatucci2009zero}, i.e.,
designing a classifier which is extensible in the output space.

\cite{hsu2009multi}, motivated by advances in compressed sensing, utilized
a random embedding of the labels along with greedy sparse decoding
strategy. For the multilabel case, \cite{tai2012multilabel} construct
a low-dimensional embedding using principal components on the empirical
label covariance, which they utilize along with a greedy sparse decoding
strategy.  For multivariate regression, \cite{breiman1997predicting}
use the principal components of the empirical label covariance to define
a shrinkage estimator which exploits correlations between the labels
to improve accuracy.  In these works, the motivation for embeddings
was primarily statistical benefit.  Conversely, \cite{weston2011wsabie}
motivate their ranking-loss optimized embeddings solely by computational
considerations of inference time and space complexity.

Multiple authors leverage side information about the classes, such as
a taxonomy or graph, in order to learn a label representation which is
felicitous for classification, e.g. when composed with online learning
\citep{dekel2004large}; Bayesian learning~\citep{decoro2007bayesian};
support vector machines~\citep{bennett2009refined}; and decision tree
ensembles~\citep{schietgat2010predicting}.  Our embedding approach
neither requires nor exploits such side information, and is therefore
applicable to different scenarios, but is potentially suboptimal
when side information is present.  However, our embeddings
can be complementary to such techniques when side information is
not present, as some approaches condense side information into a
similarity matrix between classes, e.g., the sub-linear inference
approach of \cite{cisse2012learning} and the large margin approach of
\cite{weinberger2009large}.  Our embeddings provide a low-rank similarity
matrix between classes in factored form, i.e., represented in $O(k c)$
rather than $O(c^2)$ space, which can be composed with these techniques.
Analogously, \cite{bengio2010label} utilize a surrogate classifier rather
than side information to define a similarity matrix between classes;
our procedure can efficiently produce a similarity matrix which can ease
the computational burden of this portion of their procedure.

Another intriguing use of side information about the classes is to enable
zero-shot learning.  To this end, several authors have exploited the
textual nature of classes in image annotation to learn an embedding over the
classes which generalizes to novel classes, e.g., \cite{frome2013devise}
and \cite{socher2013zero}.  Our embedding technique does not address
this problem.

\cite{gopal2013recursive} focus nearly exclusively on the statistical
benefit of incorporating label structure by overcoming the space and
time complexity of large-scale one-against-all classification via
distributed training and inference.  Specifically, they utilize side
information about the classes to regularize a set of one-against-all
classifiers towards each other.  This leads to state-of-the-art predictive
performance, but the resulting model has high space complexity, e.g.,
terabytes of parameters for the LSHTC~\citep{LSHTC4} dataset we utilize
in section \ref{subsec:lshtc}.  This necessitates distributed learning
and distributed inference, the latter being a more serious objection in
practice. In contrast, our embedding technique mitigates space complexity
and avoids model parallelism.

Our objective in equation \eqref{eqn:rankconstrained} is highly
related to that of partial least squares~\citep{geladi1986partial},
as Algorithm \ref{alg:rembed} corresponds to a randomized algorithm
for PLS if the features have been whitened.\footnote{More precisely,
if the feature covariance is a rotation.}  Unsurprisingly, supervised
dimensionality reduction techniques such as PLS can be much better than
unsupervised dimensionality reduction techniques such as PCA regression
in the discriminative setting if the features vary in ways irrelevant
to the classification task~\citep{barker2003partial}.

Two other classical procedures for supervised dimensionality
reduction are Fisher Linear Discriminant~\citep{rao1948} and
Canonical Correlation Analysis~\citep{hotelling1936relations}.
For multiclass problems these two techniques yield the
same result~\citep{bartlett1938further,barker2003partial},
although for multilabel problems they are distinct.  Indeed,
extension of FLD to the multilabel case is a relatively recent
development~\citep{wang2010multi} whose straightforward implementation
does not appear to be computationally viable for large number of classes.
CCA and PLS are highly related, as CCA maximizes latent correlation
and PLS maximizes latent covariance~\citep{barker2003partial}.
Furthermore, CCA produces equivalent results to PLS if the features
are whitened~\citep{sun2009equivalence}. Therefore, there is no obvious
statistical reason to prefer CCA to our proposal in this context.

Regarding computational considerations, scalable CCA algorithms
are available~\citep{lu2014large,mineiro2014randomized}, but it
remains open how to specialize them to this context to leverage the
equivalent of equation \eqref{eqn:multop}; whereas, if CCA is desired,
Algorithm \ref{alg:rembed} can be utilized in conjunction with whitening
pre-processing.

Text is one the common input domains over which large-scale multiclass
and multilabel problems are defined.  There has been substantial recent
work on text embeddings, e.g., word2vec~\citep{mikolov2013efficient},
which (empirically) provide analogous statistical and computational
benefits despite being unsupervised.  The text embedding technique
of \cite{lebret2013word} is a particularly interesting comparison
because it is a variant of Hellinger PCA which leverages sequential
information.  This suggests that unsupervised dimensionality reduction
approaches can work well when additional structure of the input domain
is incorporated, in this case by modeling word burstiness with the
square root nonlinearity~\citep{jegou2009burstiness} and word order via
decomposing neighborhood statistics.  Nonetheless \cite{lebret2013word}
note that when maximum statistical performance is desired, the embeddings
must be fine-tuned to the particular task, i.e., supervised dimensionality
reduction is required.

Another plausible regularization technique which mitigates inference
space and time complexity is L1 regularization~\citep{lokhorst1999}.
One reason to prefer low-rank regularization to L1 regularization is
if the prediction covariance of equation \eqref{eqn:darkknowledge}
is well-modeled by a low-rank matrix.

\section{Experiments} \label{sec:experiments}

\begin{algorithm}[t]
  \caption{Stagewise classification algorithm utilized for experiments.  Loss is either log loss (multiclass) or independent log loss per class (multilabel).}
  \begin{algorithmic}[1]
    \Statex

    \Function{DecoderTrain}{$k, X \in \R^{n \times d}, Y \in \R^{n \times c}, \phi$}
      \State $(R,\sim) \leftarrow \mathrm{\RandEmbed}(k, X, Y)$ \Comment{or other comparison embedding}
      \State $W^* \leftarrow \argmin_W \| Y R - X W \|_F^2$
      \State $\hat Y \leftarrow \phi(X W^*)$ \Comment{$\phi$ is an optional random feature map} \label{line:featuremap}
      \State $Q^* \leftarrow \argmin_Q \mathrm{loss} (Y, \hat Y Q)$ \Comment{early-stopped, see text} \label{line:lossstep}
      \State \Return $(W^*,  Q^*)$
    \EndFunction
  \end{algorithmic}
  \label{alg:logfit}
\end{algorithm}

The goal of these experiments is to demonstrate the computational
viability and statistical benefits of the embedding algorithm, not to
advocate for a particular classification algorithm per se.  We
utilize classification tasks for demonstration, and utilize our
embedding strategy as part of algorithm \ref{alg:logfit}, but focus 
our attention on the impact of the embedding on the result.

\begin{table}[h]
\centering
\caption{Data sets used for experimentation and times to compute an embedding.}
\begin{tabular}{|c|c|c|c|c|c|c|c|} \hline
Dataset & Type & Modality & Examples & Features & Classes & \multicolumn{2}{c|}{\RandEmbed} \\ 
& & & & & & $k$ & Time (sec) \\ \hline
ALOI & Multiclass & Vision & 108K & 128 & 1000 & 50 & 4 \\
ODP & Multiclass & Text & $\sim$ 1.5M & $\sim$ 0.5M & $\sim$ 100K & 300 & 6,530 \\
LSHTC & Multilabel & Text & $\sim$ 2.4M & $\sim$ 1.6M & $\sim$ 325K & 500 & 8,006 \\ \hline
\end{tabular}
\label{tab:datasets}
\end{table}

In table \ref{tab:datasets} we present some statistics about the datasets
we use in this section as well as times required to compute an embedding
for the dataset.  Unless otherwise indicated, all timings presented in
the experiments section are for a Matlab implementation running on a
standard desktop, which has dual 3.2Ghz Xeon E5-1650 CPU and 48Gb of RAM.

\subsection{ALOI}

ALOI is a color image collection of one-thousand small objects recorded
for scientific purposes~\citep{geusebroek2005amsterdam}. The number
of classes in this data set does not qualify as extreme by current
standards, but we begin with it as it will facilitate comparison
with techniques which in our other experiments are intractable
on a single machine.  For these experiments we will consider test
classification accuracy utilizing the same train-test split and features
from \cite{choromanska2014logarithmic}.  Specifically there is a fixed
train-test split of 90:10 for all experiments and the representation is
linear in 128 raw pixel values.

Algorithm \ref{alg:rembed} produces an embedding matrix whose transpose is
a squared-loss optimal decoder.  In practice, optimizing the decode matrix
for logistic loss as described in Algorithm \ref{alg:logfit} gives much
better results.  This is by far the most computationally demanding step
in this experiment, e.g., it takes 4 seconds to compute the embedding but
300 seconds to perform the logistic regression.  Fortunately the number
of features (i.e., embedding dimensionality) for this logistic regression
is modest so the second order techniques of \cite{agarwal2013least} are
applicable (in particular, their Algorithm~1 with a simple modification
to include acceleration~\citep{nesterov1983method,beck2009fast}).
We determine the number of fit iterations for the logistic regression
by extracting a hold-out set from the training set and monitoring
held-out loss.  We do not use a random feature map, i.e., $\phi$
in line \ref{line:featuremap} of Algorithm \ref{alg:logfit} is the
identity function.

\begin{table}[h]
\centering
\caption{ALOI results.  $k=50$ for all embedding strategies.}
\begin{tabular}{|c|c|c|c|c|c|c|} \hline
Method & RE + LR & PCA + LR & CS + LR & LR & OAA & LT \\ \hline
Test Error & 9.7\% & 9.7\% & 10.8\% & 10.8\% & 11.5\% & 16.5\% \\ \hline
\end{tabular}
\label{tab:aloi}
\end{table}

We compare to several different strategies in table \ref{tab:aloi}.
OAA is the one-against-all reduction of multiclass to binary.  LR is
a standard logistic regression, i.e., learning directly from the
original features.  Both of these options are intractable on a single
machine for our other data sets.  We also compare against Lomtree (LT),
which has training and test time complexity logarithmic in the number
of classes~\citep{choromanska2014logarithmic}.  Both OAA and LT are
provided by the Vowpal Wabbit~\citep{langford2007} machine learning tool.

The remaining techniques are variants of Algorithm \ref{alg:logfit} using
different embedding strategies.  PCA + LR refers to logistic regression
after first projecting the features onto their top principal components.
CS + LR refers to logistic regression on a label embedding which is a
random Gaussian matrix suitable for compressed sensing.  Finally RE +
LR is \RandEmbed composed with logistic regression.  These techniques
were all implemented in Matlab.

Interestingly, OAA underperforms the full logistic regression.
\RandEmbed combined with logistic regression outperforms logistic
regression, suggesting a beneficial effect from low-rank regularization.
Compressed sensing is able to match the performance of the full logistic
regression while being computationally more tractable, but underperforms
\RandEmbed. Lomtree has the worst prediction performance but the lowest
computational overhead when the number of classes is large.

\begin{figure}[t]
\centering
\begin{subfigure}[t]{0.45 \textwidth}
\includegraphics*[height=1.9in]{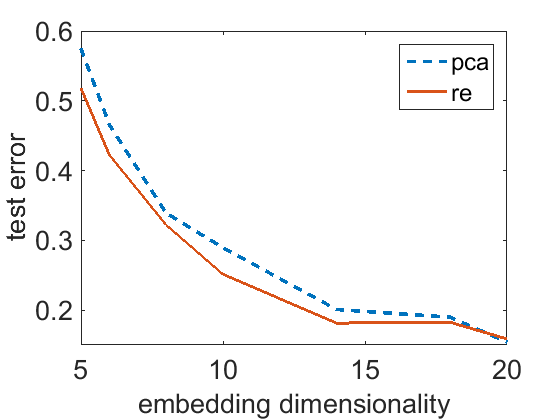}
\caption{Performance of logistic regression on ALOI when combined with either a
feature embedding (PCA) or label embedding (RE).}
\label{fig:aloipcavsre}
\end{subfigure}
\hspace{0.05 \textwidth}
\begin{subfigure}[t]{0.45 \textwidth}
\includegraphics*[height=2in,clip=true,trim=100 225 100 225]{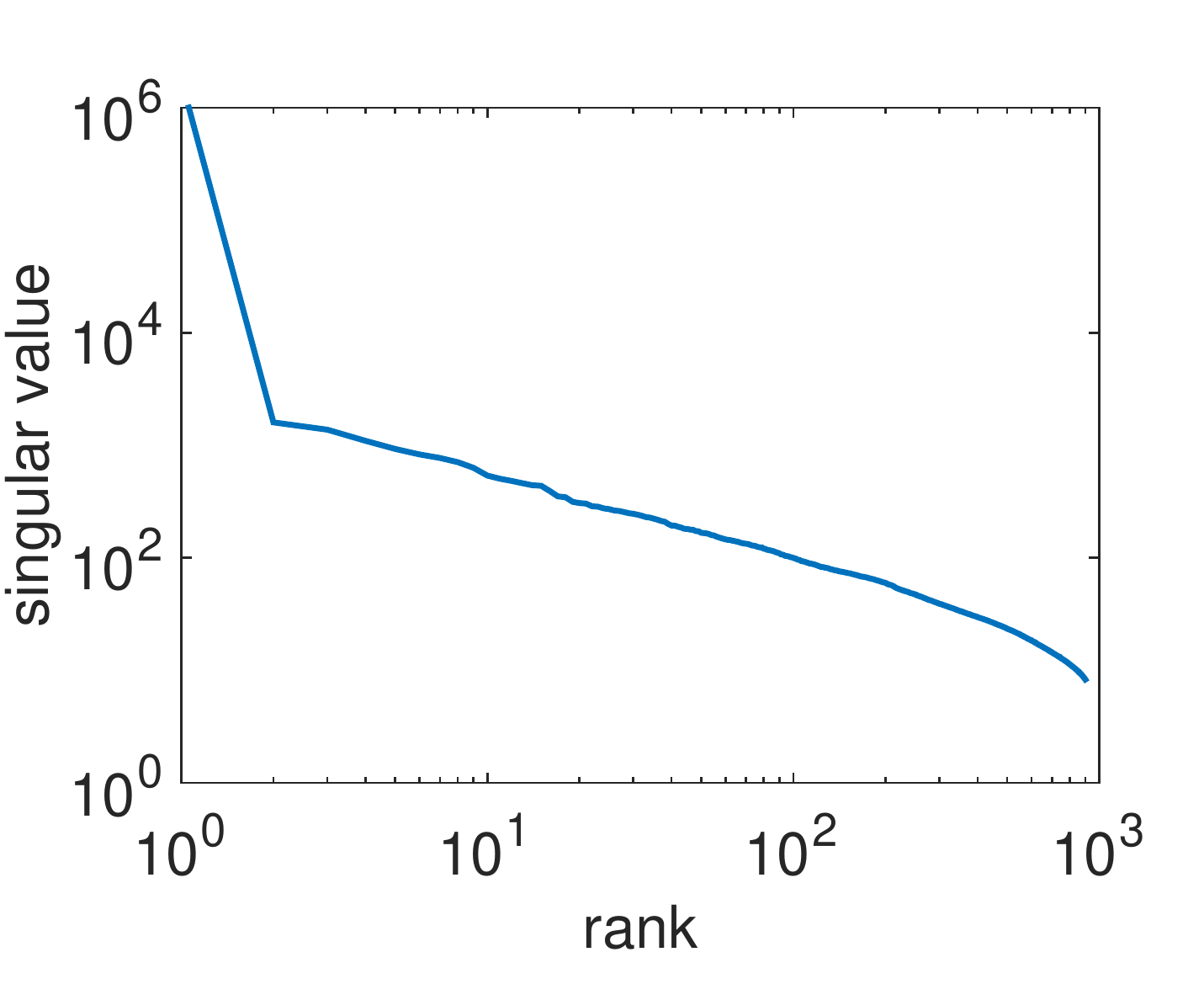}
\caption{The empirical label covariance spectrum for LSHTC.}
\label{fig:lshtcspectrum}
\end{subfigure}
\end{figure}

At $k=50$, there is no difference in quality between using the
\RandEmbed (label) embedding and the PCA (feature) embedding.  This is
not surprising considering the effective rank of the covariance matrix
of ALOI is 70.  For small embedding dimensionalities, however, PCA
underperforms \RandEmbed as indicated in Figure \ref{fig:aloipcavsre}.
For larger numbers of output classes, where the embedding dimension will
be a small fraction of the number of classes by computational necessity,
we anticipate PCA regression will not be competitive.

Note that, in addition to better statistical performance, all of the
``embedding + LR'' approaches have lower space complexity $O(k (c+d))$
than direct logistic regression $O (c d)$.  For ALOI the savings are
modest (255600 bytes vs. 516000 bytes) because the input dimensionality is
only $d=128$, but for larger problems the space savings are necessary for
feasible implementation on a single commodity computer.  Inference time
on ALOI is identical for embedding and direct approaches in practice
(both achieving $\approx 170$k examples/sec).

\subsection{ODP}

The Open Directory Project \citep{ODP} is a public human-edited
directory of the web which was processed by \cite{bennett2009refined}
into a multiclass data set.  For these experiments we will consider test
classification error utilizing the same train-test split, features,
and labels from \cite{choromanska2014logarithmic}.  Specifically there is
a fixed train-test split of 2:1 for all experiments, the representation
of document is a bag of words, and the unique class assignment for each
document is the most specific category associated with the document.

The procedures are the same as in the previous experiment, except that we
do not compare to OAA or full logistic regression due to intractability
on a single machine.

\begin{table}[t]
\centering
\caption{ODP results.  $k=300$ for all embedding strategies.}
\begin{tabular}{|c|c|c|c|c|} \hline
Method & RE + LR & CS + LR & PCA + LR & LT \\ \hline
Test Error & 83.15\% & 85.14\% & 90.37\% & 93.46\% \\ \hline
\end{tabular}
\label{tab:odp}
\end{table}

The combination of \RandEmbed and logistic regression result is, to
the best of our knowledge, the best published result on this dataset.
PCA logistic regression has a performance gap compared to \RandEmbed and
logistic regression.  The poor performance of PCA logistic regression
is doubly unsurprising, both for general reasons previously discussed,
and due to the fact that covariance matrices of text data typically have
a long plateau of weak spectral decay. In other words, for text problems
projection dimensions quickly become nearly equivalent in terms of input
reconstruction error, and common words and word combinations are not
discriminative.  In contrast, \RandEmbed leverages the spectral properties
of the prediction covariance of equation \eqref{eqn:darkknowledge},
rather than the spectral properties of the input features.

Finally, we remark the following: although inference (i.e., finding the
maximum output) is linear in the number of classes, the constant factors
are favorable due to modern vectorized processors, and therefore proceeds
at $\approx 1700$ examples/sec for the embedding based approaches.

\subsection{LSHTC} \label{subsec:lshtc}

The Large Scale Hierarchical Text Classification Challenge (version 4)
was a public competition involving multilabel classification of documents
into approximately 300,000 categories \citep{LSHTC4}. The training and
test files are available from the Kaggle platform.  The features are
bag of words representations of each document.

\subsubsection{Embedding Quality Assessment}

\begin{table}[!h]
\centering
\caption{Embedding quality for LSHTC.  $k=100$ for all embedding strategies.}
\begin{tabular}{|c|c|c|c|c|} \hline
Method & Most fraternal & CS & PLST & \RandEmbed \\ \hline
Sibling Fraction & 0.32\% & 3.08\% & 19.65\% & 23.61\% \\ \hline
\end{tabular}
\label{tab:lshtcdag}
\end{table}

A representation of a DAG hierarchy associated with the classes is also
available.  We used this to assess the quality of various embedding
strategies independent of classification performance.  In particular,
we computed the fraction of class embeddings whose nearest neighbor
was also a sibling in the DAG, as shown in Table \ref{tab:lshtcdag}.
``Most fraternal'' refers to an embedding which arranges for every
category's nearest neighbor in the embedding to be the node in the
DAG with the most siblings, i.e., the constant predictor baseline for
this task.  PLST~\citep{tai2012multilabel} has performance close to
\RandEmbed according to this metric, so the 3.2 average nonzero classes
per example is apparently enough for the approximation underlying PLST
to be reasonable.

\subsubsection{Empirical Label Covariance Spectrum}

Our embedding approach is based upon a low-rank assumption
for the (unobservable) prediction covariance of equation
\eqref{eqn:darkknowledge}.  Because LSHTC is a multi-label dataset,
we can use the empirical label covariance as a proxy to investigate the
spectral properties of the prediction covariance and test our assumption.
We used Algorithm \ref{alg:rpca} (i.e., two pass randomized PCA) to
estimate the spectrum of the empirical label covariance, shown in Figure
\ref{fig:lshtcspectrum}.  The spectrum decays modestly and suggests that
an embedding dimension of $k \approx 1000$ or more might be necessary
for good classification performance.

\subsubsection{Classification Performance}

We built an end-to-end classifier using an approximate kernelized
variant of Algorithm \ref{alg:logfit}, where we processed the embeddings
with Random Fourier Features~\citep{rahimi2007random}, i.e., in line
\ref{line:featuremap} of Algorithm \ref{alg:logfit} we use a random cosine
feature map for $\phi$.  We found Cauchy distributed random vectors,
corresponding to the Laplacian kernel, gave good results.  We used 4,000
random features and tuned the kernel bandwidth via cross-validation on
the training set.

The LSHTC competition metric is macro-averaged F1, which emphasizes
performance on rare classes.  However, we are using a multilabel
classification algorithm which maximizes accuracy of predictions without
regard to the importance of rare classes.  Therefore we compare with
published results of \cite{prabhu2014fastxml}, who report example-averaged
precision-at-$k$ on the label ordering induced for each example.
To facilitate comparison we do a 75:25 train-test split of the public
training set, which is the same proportions as in their experiments
(albeit a different split).

Based upon the previous spectral analysis, we anticipate a large embedding
dimension is required for best results.  With our current implementation, 
up to the limit of available memory in our desktop machine ($k=800$)
we found increasing embedding dimensionality improved performance.

\begin{table}[t]
\centering
\caption{LSHTC results.}
\begin{tabular}{|c|c|c|c|c|} \hline
Method &  RE ($k=800$) + ILR & RE ($k=500$) + ILR &  FastXML & LPSR-NB \\ \hline
Precision-at-1& 53.39\% & 52.84\% & 49.78\% & 27.91\% \\ \hline 
\end{tabular}
\label{tab:lshtc}
\end{table}

``RE $(k = \ldots)$ + ILR'' corresponds for $\RandEmbed$ coupled with
independent (kernel) logistic regression, i.e., Algorithm \ref{alg:logfit}.
LPSR-NB is the Label Partitioning by Sub-linear Ranking algorithm of
\cite{weston2013label} composed with a Naive Bayes base learner, as
reported in \cite{prabhu2014fastxml}, where they also introduce and
report precision for the multilabel tree learning algorithm FastXML.
Inference for our best model proceeds at $\approx$ 60 examples/sec,
substantially slower than for ODP, due to the larger output space,
larger embedding dimensionality, and the use of random Fourier features.

\section{Discussion} \label{sec:discussion}

In this paper we identify a correspondence between rank constrained
regression and label embedding, and we exploit that correspondence along
with randomized matrix decomposition techniques to develop a fast label
embedding algorithm.

To facilitate analysis and implementation, we focused on linear
prediction, which is equivalent to a simple neural network architecture
with a single linear hidden layer bottleneck.  Because linear predictors
perform well for text classification, we obtained excellent
experimental results, but more sophistication is required for tasks
where deep architectures are state-of-the-art.  Although the analysis
presented herein would not strictly be applicable, it is plausible that
replacing line \ref{lin:learningstep} in Algorithm \ref{alg:rembed}
with an optimization over a deep architecture could yield good embeddings.
This would be computationally beneficial as reducing the number of
outputs (i.e., predicting embeddings rather than labels) would mitigate
space constraints for GPU training.

Our technique leverages the (putative) low-rank structure of the
prediction covariance of equation \eqref{eqn:darkknowledge}.  For some
problems a low-rank plus sparse assumption might be more appropriate.
In such cases combining our technique with L1 regularization, e.g., on
a classification residual or on separately regularized direct 
connections from the original inputs, might yield superior results.

\subsubsection*{Acknowledgments}

We thank John Langford for providing the ALOI and ODP data sets.

\bibliography{rembed}

\begin{thebibliography}{10}
\providecommand{\url}[1]{\texttt{#1}}
\providecommand{\urlprefix}{URL }

\bibitem{agarwal2013least}
Agarwal, A., Kakade, S.M., Karampatziakis, N., Song, L., Valiant, G.: Least
  squares revisited: Scalable approaches for multi-class prediction. In:
  Proceedings of The 31st International Conference on Machine Learning. pp.
  541--549 (2014)

\bibitem{barker2003partial}
Barker, M., Rayens, W.: Partial least squares for discrimination. Journal of
  chemometrics  17(3),  166--173 (2003)

\bibitem{bartlett1938further}
Bartlett, M.S.: Further aspects of the theory of multiple regression. In:
  Mathematical Proceedings of the Cambridge Philosophical Society. vol.~34, pp.
  33--40. Cambridge Univ Press (1938)

\bibitem{beck2009fast}
Beck, A., Teboulle, M.: A fast iterative shrinkage-thresholding algorithm for
  linear inverse problems. SIAM Journal on Imaging Sciences  2(1),  183--202
  (2009)

\bibitem{bengio2010label}
Bengio, S., Weston, J., Grangier, D.: Label embedding trees for large
  multi-class tasks. In: Advances in Neural Information Processing Systems. pp.
  163--171 (2010)

\bibitem{bennett2009refined}
Bennett, P.N., Nguyen, N.: Refined experts: improving classification in large
  taxonomies. In: Proceedings of the 32nd international ACM SIGIR conference on
  Research and development in information retrieval. pp. 11--18. ACM (2009)

\bibitem{breiman1997predicting}
Breiman, L., Friedman, J.H.: Predicting multivariate responses in multiple
  linear regression. Journal of the Royal Statistical Society: Series B
  (Statistical Methodology)  59(1),  3--54 (1997)

\bibitem{choromanska2014logarithmic}
Choromanska, A., Langford, J.: Logarithmic time online multiclass prediction.
  arXiv preprint arXiv:1406.1822  (2014)

\bibitem{cisse2012learning}
Ciss{\'e}, M., Arti{\`e}res, T., Gallinari, P.: Learning compact class codes
  for fast inference in large multi class classification. In: Machine Learning
  and Knowledge Discovery in Databases, pp. 506--520. Springer (2012)

\bibitem{decoro2007bayesian}
DeCoro, C., Barutcuoglu, Z., Fiebrink, R.: Bayesian aggregation for
  hierarchical genre classification. In: ISMIR. pp. 77--80 (2007)

\bibitem{dekel2004large}
Dekel, O., Keshet, J., Singer, Y.: Large margin hierarchical classification.
  In: Proceedings of the twenty-first international conference on Machine
  learning. p.~27. ACM (2004)

\bibitem{deng2009imagenet}
Deng, J., Dong, W., Socher, R., Li, L.J., Li, K., Fei-Fei, L.: Imagenet: A
  large-scale hierarchical image database. In: Computer Vision and Pattern
  Recognition, 2009. CVPR 2009. IEEE Conference on. pp. 248--255. IEEE (2009)

\bibitem{ODP}
{DMOZ}: The open directory project (2014), \url{http://dmoz.org/}

\bibitem{friedland2007generalized}
Friedland, S., Torokhti, A.: Generalized rank-constrained matrix
  approximations. SIAM Journal on Matrix Analysis and Applications  29(2),
  656--659 (2007)

\bibitem{frome2013devise}
Frome, A., Corrado, G.S., Shlens, J., Bengio, S., Dean, J., Mikolov, T.,
  et~al.: Devise: A deep visual-semantic embedding model. In: Advances in
  Neural Information Processing Systems. pp. 2121--2129 (2013)

\bibitem{geladi1986partial}
Geladi, P., Kowalski, B.R.: Partial least-squares regression: a tutorial.
  Analytica chimica acta  185,  1--17 (1986)

\bibitem{geusebroek2005amsterdam}
Geusebroek, J.M., Burghouts, G.J., Smeulders, A.W.: The {A}msterdam library of
  object images. International Journal of Computer Vision  61(1),  103--112
  (2005)

\bibitem{gopal2013recursive}
Gopal, S., Yang, Y.: Recursive regularization for large-scale classification
  with hierarchical and graphical dependencies. In: Proceedings of the 19th ACM
  SIGKDD international conference on Knowledge discovery and data mining. pp.
  257--265. ACM (2013)

\bibitem{halko2011finding}
Halko, N., Martinsson, P.G., Tropp, J.A.: Finding structure with randomness:
  Probabilistic algorithms for constructing approximate matrix decompositions.
  SIAM review  53(2),  217--288 (2011)

\bibitem{hotelling1936relations}
Hotelling, H.: Relations between two sets of variates. Biometrika pp. 321--377
  (1936)

\bibitem{hsu2009multi}
Hsu, D., Kakade, S., Langford, J., Zhang, T.: Multi-label prediction via
  compressed sensing. In: NIPS. vol.~22, pp. 772--780 (2009)

\bibitem{izenman1975reduced}
Izenman, A.J.: Reduced-rank regression for the multivariate linear model.
  Journal of multivariate analysis  5(2),  248--264 (1975)

\bibitem{jegou2009burstiness}
J{\'e}gou, H., Douze, M., Schmid, C.: On the burstiness of visual elements. In:
  Computer Vision and Pattern Recognition, 2009. CVPR 2009. IEEE Conference on.
  pp. 1169--1176. IEEE (2009)

\bibitem{LSHTC4}
Kaggle: Large scale hierarchical text classification (2014),
  \url{http://www.kaggle.com/c/lshtc}

\bibitem{kosmopoulos2010ecir}
Kosmopoulos, A., Gaussier, E., Paliouras, G., Aseervatham, S.: The {ECIR} 2010
  large scale hierarchical classification workshop. In: ACM SIGIR Forum.
  vol.~44, pp. 23--32. ACM (2010)

\bibitem{langford2007}
Langford, J.: {V}owpal {W}abbit.
  \url{https://github.com/JohnLangford/vowpal_wabbit/wiki} (2007)

\bibitem{lebret2013word}
Lebret, R., Collobert, R.: Word emdeddings through hellinger pca. arXiv
  preprint arXiv:1312.5542  (2013)

\bibitem{liberty2013simple}
Liberty, E.: Simple and deterministic matrix sketching. In: Proceedings of the
  19th ACM SIGKDD international conference on Knowledge discovery and data
  mining. pp. 581--588. ACM (2013)

\bibitem{lokhorst1999}
Lokhorst, J.: The lasso and generalised linear models. Tech. rep., University
  of Adelaide, Adelaide (1999)

\bibitem{lu2014large}
Lu, Y., Foster, D.P.: Large scale canonical correlation analysis with iterative
  least squares. arXiv preprint arXiv:1407.4508  (2014)

\bibitem{mikolov2013efficient}
Mikolov, T., Chen, K., Corrado, G., Dean, J.: Efficient estimation of word
  representations in vector space. arXiv preprint arXiv:1301.3781  (2013)

\bibitem{mineirogit2015}
Mineiro, P.: randembed. \url{https://github.com/pmineiro/randembed} (2015)

\bibitem{mineiro2014randomized}
Mineiro, P., Karampatziakis, N.: A randomized algorithm for {CCA}. arXiv
  preprint arXiv:1411.3409  (2014)

\bibitem{nesterov1983method}
Nesterov, Y.: A method of solving a convex programming problem with convergence
  rate ${O}(1/k^2)$. Dokl. Akad. Nauk SSSR  269,  543--547 (1983)

\bibitem{palatucci2009zero}
Palatucci, M., Pomerleau, D., Hinton, G.E., Mitchell, T.M.: Zero-shot learning
  with semantic output codes. In: Advances in neural information processing
  systems. pp. 1410--1418 (2009)

\bibitem{prabhu2014fastxml}
Prabhu, Y., Varma, M.: Fastxml: a fast, accurate and stable tree-classifier for
  extreme multi-label learning. In: Proceedings of the 20th ACM SIGKDD
  international conference on Knowledge discovery and data mining. pp.
  263--272. ACM (2014)

\bibitem{rahimi2007random}
Rahimi, A., Recht, B.: Random features for large-scale kernel machines. In:
  Advances in neural information processing systems. pp. 1177--1184 (2007)

\bibitem{rao1948}
Rao, C.R.: The utilization of multiple measurements in problems of biological
  classification. Journal of the Royal Statistical Society. Series B
  (Methodological)  10(2),  pp. 159--203 (1948),
  \url{http://www.jstor.org/stable/2983775}

\bibitem{schietgat2010predicting}
Schietgat, L., Vens, C., Struyf, J., Blockeel, H., Kocev, D., D{\v{z}}eroski,
  S.: Predicting gene function using hierarchical multi-label decision tree
  ensembles. BMC Bioinformatics  11, ~2 (2010)

\bibitem{socher2013zero}
Socher, R., Ganjoo, M., Manning, C.D., Ng, A.: Zero-shot learning through
  cross-modal transfer. In: Advances in Neural Information Processing Systems.
  pp. 935--943 (2013)

\bibitem{sun2009equivalence}
Sun, L., Ji, S., Yu, S., Ye, J.: On the equivalence between canonical
  correlation analysis and orthonormalized partial least squares. In: IJCAI.
  vol.~9, pp. 1230--1235 (2009)

\bibitem{tai2012multilabel}
Tai, F., Lin, H.T.: Multilabel classification with principal label space
  transformation. Neural Computation  24(9),  2508--2542 (2012)

\bibitem{wang2010multi}
Wang, H., Ding, C., Huang, H.: Multi-label linear discriminant analysis. In:
  Computer Vision--ECCV 2010, pp. 126--139. Springer (2010)

\bibitem{weinberger2009large}
Weinberger, K.Q., Chapelle, O.: Large margin taxonomy embedding for document
  categorization. In: Advances in Neural Information Processing Systems. pp.
  1737--1744 (2009)

\bibitem{weston2011wsabie}
Weston, J., Bengio, S., Usunier, N.: Wsabie: Scaling up to large vocabulary
  image annotation. In: IJCAI. vol.~11, pp. 2764--2770 (2011)

\bibitem{weston2013label}
Weston, J., Makadia, A., Yee, H.: Label partitioning for sublinear ranking. In:
  Proceedings of the 30th International Conference on Machine Learning
  (ICML-13). pp. 181--189 (2013)

\end{thebibliography}
\bibliographystyle{splncs03}

\end{document}